\documentclass[lettersize,journal]{IEEEtran}
\usepackage{amsmath,amsfonts}
\usepackage{array}
\usepackage{textcomp}
\usepackage{stfloats}
\usepackage{url}
\usepackage{verbatim}
\usepackage{graphicx}
\usepackage{cite}

\usepackage{amssymb}
\usepackage{booktabs}
\usepackage{bm} 
\usepackage{multirow}
\usepackage[table]{xcolor}
\usepackage{subcaption}

\usepackage{pifont}
\newcommand{\xmark}{\ding{55}}%
\newcommand{\cmark}{\ding{51}}%

\newcommand{\mr}[1]{\textcolor{black}{#1}}

\usepackage{algorithm2e}
\RestyleAlgo{ruled}
\LinesNumbered
\SetKwInput{KwData}{Require}

\begin{document}

\title{Admitting Ignorance Helps the Video Question Answering Models to Answer}

\author{Haopeng Li, Tom Drummond, Mingming Gong, Mohammed Bennamoun, and Qiuhong Ke
        % <-this % stops a space
\thanks{Haopeng Li and Tom Drummond are with the School of Computing and Information Systems, University of Melbourne.  E-mail: haopeng.li@student.unimelb.edu.au, tom.drummond@unimelb.edu.au.}% <-this % stops a space
\thanks{Mingming Gong is with the School of Mathematics and Statistics, University of Melbourne. E-mail: mingming.gong@unimelb.edu.au.}
\thanks{Mohammed Bennamoun is with the school of Physics, Maths and Computing, The University of Western Australia. E-mail: mohammed.bennamoun@uwa.edu.au.}
\thanks{Qiuhong Ke is with the Department of Data Science \& AI, Monash University.  E-mail: qiuhong.ke@monash.edu.}
\thanks{This research was partially supported by the Australian Government through the Australian Research Council's DECRA funding scheme (Grant No.: DE250100030) and DP funding scheme (Grant No.: DP210101682).}
\thanks{Corresponding Author: Qiuhong Ke.}
}

\markboth{Journal of \LaTeX\ Class Files,~Vol.~14, No.~8, August~2021}%
{Shell \MakeLowercase{\textit{et al.}}: A Sample Article Using IEEEtran.cls for IEEE Journals}


\maketitle

\begin{abstract}
Significant progress has been made in the field of video question answering (VideoQA) thanks to deep learning and large-scale pretraining. Despite the presence of sophisticated model structures and powerful video-text foundation models, most existing methods focus solely on maximizing the correlation between answers and video-question pairs during training. We argue that these models often establish shortcuts, resulting in spurious correlations between questions and answers, especially when the alignment between video and text data is suboptimal. To address these spurious correlations, we propose a novel training framework in which the model is compelled to acknowledge its ignorance when presented with an intervened question, rather than making guesses solely based on superficial question-answer correlations. We introduce methodologies for intervening in questions, utilizing techniques such as displacement and perturbation, and design frameworks for the model to admit its lack of knowledge in both multi-choice VideoQA and open-ended settings. In practice, we integrate a state-of-the-art model into our framework to validate its effectiveness. The results clearly demonstrate that our framework can significantly enhance the performance of VideoQA models with minimal structural modifications.
\end{abstract}


\section{Introduction}

\IEEEPARstart{V}{ideo} Question Answering (VideoQA) has experienced notable progressions, particularly in the realm of deep learning techniques. These contributions can be broadly categorized into two areas: 1) proposing sophisticated model structures to address specific challenges in VideoQA, such as the use of bi-linear attention mechanisms \cite{kim2018bilinear,seo2021attend} for video-text alignment or the implementation of conditional graph hierarchies for multi-granular understanding of linguistic concepts \cite{xiao2022video}; 2) pretraining foundation models on large-scale data to enhance generalization abilities, followed by fine-tuning for various downstream tasks \cite{zellers2021merlot,fu2021violet,zeng2022x,wang2022internvideo,wang2023all}. Despite the progress made through novel model structures and large-scale video-text model pretraining, most of these approaches share a common objective in VideoQA: maximizing the correlation between the answer and the video-question pair. However, this goal has limitations, as it can lead to a dilemma where the model must decide whether to trust the video or the question, especially when video-text alignment is suboptimal. In such cases, the model tends to rely solely on either the video or the question for answer prediction, as modeling the correlation between the answer and the video or question alone is more straightforward.

\begin{figure}[t]
    \centering
    \includegraphics[width=\linewidth]{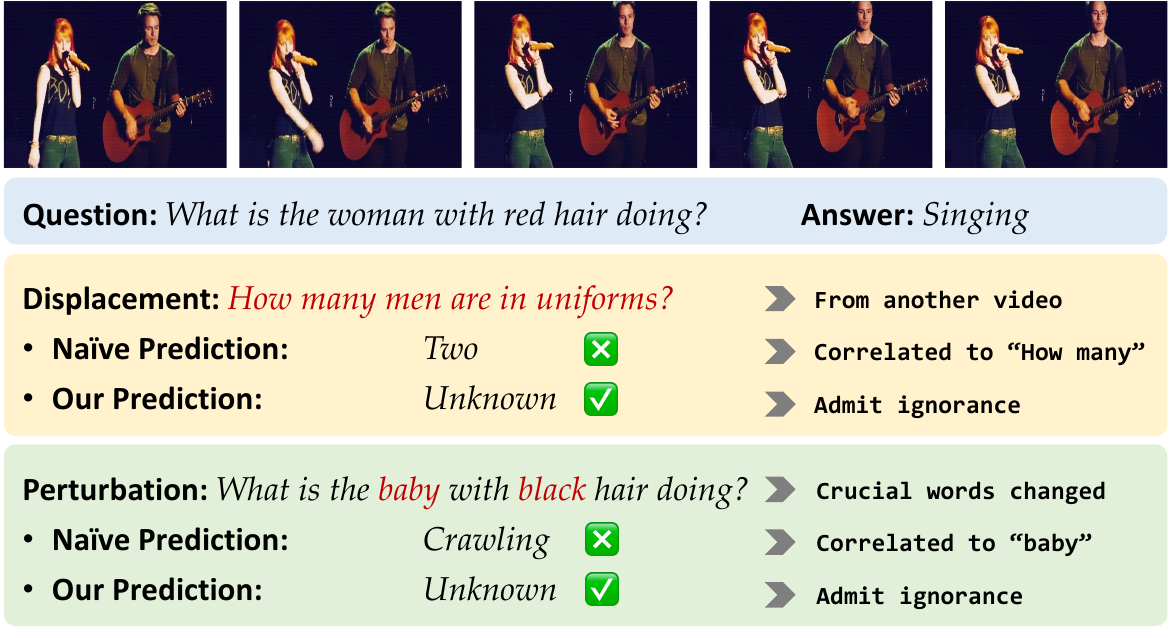}
    \caption{
    The difference between conventional VideoQA and our ignorance-admitting VideoQA lies in how they handle spurious correlations. In existing VideoQA, when video-question alignment fails, the model often resorts to guessing the answer based on spurious correlations between the question and the answer. In contrast, our framework disrupts such correlations by introducing interventions (displacement and perturbation) to the questions and compelling the model to acknowledge its ignorance in response to the intervened inputs.
    }
    \label{exam1}

% \vspace{-1.5em}
\end{figure}

Such a limitation has been noted in \cite{li2022invariant,li2022equivariant,li2023transformer,li2023discovering}, which argue that simply minimizing the empirical error can lead to spurious correlations between videos and answers. By introducing interventions to the videos, these studies break such correlations and addresses the problem of answer prediction relying solely on the videos. Despite this inspiring perspective, they overlook the issue that spurious question-answer correlations  still exist, which is easier for models to identify since unimodal (question-answer) correlations are easier to capture than multimodal (video-answer) ones \cite{li2021adversarial,zang2023discovering,niu2021counterfactual,dancette2021beyond,kervadec2021roses,sheng2021human}. Within this research, we aim to break the spurious correlations between questions and answers and propose a training framework that compels the model to capture the causal relations.

In our framework, we make interventions to the questions and force the VideoQA model to admit its ignorance regarding semantically inconsistent video-question pairs, as illustrated in Fig. \ref{exam1}. By explicitly modeling the ignorance of the model, we expect that it will learn better video-question alignment and develop robust multimodal representations. In other words, we compel the model to learn the semantic correspondences between the question and the video and to acknowledge its ignorance when it tends to rely on the remembered correlations between the question and answer for answer prediction.

Specifically, we introduce two types of interventions for questions: displacement and perturbation. Displacement involves replacing the question in a video-question pair with questions from other pairs, while perturbation modifies crucial words necessary to find the answer in the question. These strategies are designed to help the model learn both global (easy) correspondences between the question and the video and local (hard) ones. We also tailor our approach to different types of VideoQA tasks, including open-ended VideoQA and multi-choice VideoQA, based on their specific formulations. Additionally, we propose a curriculum learning \cite{wang2021survey,soviany2022curriculum} strategy for diminishing ignorance. This strategy gradually trains the model, starting with the easier task of admitting ignorance when presented with intervened or original video-question pairs and progressing to the harder task of providing correct answers for the original video-question pairs. In practice, our framework is model-agnostic, allowing us to integrate state-of-the-art models to validate its effectiveness.

Our contributions are summarized as follows,

\begin{enumerate}
    \item We address the dilemma that current VideoQA formulations face and propose to break the spurious correlations between questions and answers to achieve better video-text alignment and robust multimodal representations.
    \item We introduce a novel training framework in which questions are intervened, and the model is required to admit its ignorance in response to the intervened input. Additionally, we propose an ignorance-diminishing curriculum learning strategy to balance the learning process.
    \item We apply our model-agnostic framework to existing VideoQA models and demonstrate its effectiveness. The results indicate that it can significantly enhance performance with minimal modifications to the model.
\end{enumerate}

\section{Related Work}

\subsection{Video Question Answering} 
Advanced VideoQA methods have been proposed based on deep neural networks to address the problems that exist in this task \cite{jang2017tgif,seo2021attend,xiao2022video,buch2022revisiting,li2024answering,liu2021question,yin2019memory}, which is the extension of single-image visual question answering \cite{cao2021knowledge,ma2021multitask,guo2021bilinear,cao2022bilateral,wang2024bridging,zhang2024latent,zheng2023webly}. For example, a dual-LSTM-based approach with both spatial and temporal attention is proposed in \cite{jang2017tgif}. 
MASN \cite{seo2021attend} models each object as a graph node and captures the spatial and temporal dependencies of all objects with graph neural networks. 
\mr{Besides, \cite{zhu2017uncovering} presents an approach for video question answering based on temporal structure modeling. An unsupervised encoder-decoder model is used for visual context learning, and a dual-channel ranking loss is proposed for answering questions.}
HQGA \cite{xiao2022video} is developed to model the video as a conditional graph hierarchy to align with the multi-granular nature of questions, achieving remarkable results on MSVD and MSRVTT \cite{xu2017video}. 
The atemporal probe (ATP) \cite{buch2022revisiting} is presented to degrade the video-language task to image-level understanding, providing a stronger baseline for image-level understanding in the video-language setting than random frames.
In summary, these methods focus on adapting various techniques, such as the attention mechanism \cite{jiang2020divide,jang2019video}, graph neural networks \cite{seo2021attend,wang2021dualvgr,park2021bridge}, memory networks \cite{gao2018motion,fan2019heterogeneous}, and hierarchical structures \cite{le2020hierarchical,xiao2022video}, for improved performance.

\subsection{Video-Text Pretraining}
In addition to designing sophisticated network structures for VideoQA, significant efforts have been dedicated to harnessing large-scale video-text pretraining to tackle this task \cite{zellers2021merlot,fu2021violet,zeng2022x,wang2022internvideo,wang2023all}. 
The general process in most of these works involves two main steps: 1) pretraining the video-text model using extensive data through self-supervised learning methods like contrastive learning; 2) fine-tuning the model for specific downstream tasks. 
For example, 
VIOLET presents an end-to-end video-language Transformer to model the temporal dynamics of videos, utilizing masked visual-token modeling to enhance video representation \cite{fu2021violet}. MERLOT is introduced in \cite{zellers2021merlot} to model multimodal script knowledge, leveraging millions of YouTube videos with transcribed speech. 
An All-in-one Transformer, proposed in \cite{wang2023all}, offers a unified backbone architecture capable of learning representations for both video-text multimodal data and unimodal data. 
X$^2$-VLM introduces a pre-trained video-language model that performs multi-grained vision-language pretraining, suitable for various tasks involving both images and videos \cite{zeng2022x}. 
InternVideo \cite{wang2022internvideo}, a general video-text foundation model, is designed using generative and discriminative self-supervised learning techniques. 
The pretrained model has achieved state-of-the-art performance on 39 video datasets, spanning tasks such as video recognition and video question answering. 
To further enhance fine-grained visual-text alignment in pretraining, some works generate hard negative examples on the text side \cite{momeni2023verbs,yuksekgonul2022and,doveh2023teaching}. For instance, \cite{momeni2023verbs} replaces only the verbs in the captions to encourage better verb reasoning. Similarly, in the image domain, \cite{doveh2023teaching} manipulates the textual part of paired image-text data based on language structure understanding. These visual-text pretraining methods bring about significant improvements in refining multimodal representations, inspiring us to adopt similar strategies for VideoQA.

\subsection{Debiasing in Visual Question Answering}
The biases in visual-answer relationships have been studied by \cite{dancette2021beyond,li2022invariant,li2022equivariant,li2023transformer,li2023discovering}, which mitigate the spurious video-answer correlations by causal inference. 
For example, \cite{li2023discovering} considers spatial and temporal visual cues that are question-critical, discovered by a differentiable and adaptive selection module. Alongside visual-answer biases, question-answer biases are also noticed \cite{li2021adversarial,cadene2019rubi,agrawal2018don,zang2023discovering,niu2021counterfactual,dancette2021beyond,ramakrishnan2018overcoming,kervadec2021roses,sheng2021human}. For instance, MCR \cite{zang2023discovering} addresses this issue by intervening in answers. While sharing similar motivations and aims, our method has a distinct advantage compared to existing question-answer debiasing models: our approach enhances training by manipulating only the data, making it a model-agnostic framework. In contrast, existing methods include extra modules or parallel structures, complicating its generalization to large vision-language models.

\subsection{Selective Prediction and Reliability} 
Selective prediction allows models to abstain from answering and, consequently, avoid incorrect predictions. This approach has been explored in various fields to enhance model reliability \cite{whitehead2022reliable, corbiere2019addressing, thulasidasan2019combating, xin2021art, el2010foundations, varshney2022investigating, geifman2019selectivenet, kamath2020selective}. For instance, in NLP tasks, selective prediction is integrated by adding a selector/calibrator on top of the base models \cite{varshney2022investigating,kamath2020selective}. Additionally, different selectors are evaluated for visual question answering on in-distribution data in \cite{whitehead2022reliable}, aiming to find a trade-off between model coverage and reliability. \textit{Although our method also compels models to abstain from answering, our focus is on learning better visual-text alignment for higher testing accuracy, rather than improving the reliability of models. This is achieved through the proposed ignorance-diminishing curriculum learning framework.}

\section{Revisit of VideoQA}
\label{vqaf}

Two widely-studied forms of VideoQA includes open-ended VideoQA (OEQA) and multi-choice VideoQA (MCQA). We elaborate on the formulations of each type of VideoQA as follows.

\subsection{Open-Ended VideoQA (OEQA)} 

OEQA regards VideoQA as a multi-class classification problem\footnote{\mr{Although open-ended answers can vary in length from a single word to a full sentence, many commonly used OEQA datasets simplify the task by accepting individual words as answers, and our paper adheres to this traditional, simplified approach for modeling OEQA.}}, where the answers are considered as classes, and the models are required to choose from a large answer pool given a video and a question. Concretely, given the video $V$, the question $Q$ and the answer pool $\mathcal{A}$, the model aims to predict the conditional distribution of answers $a$, i.e., 
\begin{equation}
    p(a|V,Q)=\mathrm{softmax}(f_{\theta}(V,Q)),
\end{equation}
where $f_{\theta}$ is the model parameterized by $\theta$.
The final prediction is the answer of the highest predicted probability, i.e., 
\begin{equation}
    \hat{a}=\mathop{\arg \max}_{a\in \mathcal{A}}p(a|V,Q).
\end{equation}
The cross-entropy loss is used for optimization, i.e.,
\begin{equation}
    \mathcal{L}=-\log p_{a^*},
\end{equation}
where $p_{a^*}=p(a=a^*|V,Q)$, and $a^*$ is the correct answer. 

\subsection{Multi-Choice VideoQA (MCQA)} 
Regarding MCQA, the models choose the answer from several options (e.g., five words/phrases/sentences) given a video and a question. Note that the options are different for different videos and questions. A typical way to address the multi-choice task is to first combine the question and each option as a whole, and then predict the score of correctness for each option conditioned on the video. Rigorously, given the video $V$, the question $Q$, and $N$ options $\mathcal{A}=\{a_i\}_{i=1}^N$, the model $f_{\theta}$ predicts the correctness score for each option conditioned on the video-question pair, i.e.,
\begin{equation}
    s(a_i)=f_{\theta}(V,Q,a_i),a_i\in \mathcal{A}.
\end{equation}
The predicted answer is the option of the highest score, i.e.,  
\begin{equation}
    \hat{a}=\mathop{\arg \max}_{a_i\in \mathcal{A}}s(a_i).
\end{equation}
The cross-entropy loss is applied to encourage the model to predict a higher score for the correct option, i.e.,
\begin{equation}
    \mathcal{L}=-\log \frac{e^{s(a_{i^*})}}{\sum_{j=1}^Ne^{s(a_j)}},
\end{equation}
where $i^*$ is the index of the correct option. 

\subsection{General Model Structure for VideoQA}

A typical VideoQA model usually consists of the following modules: a video encoder ($E_V$) that encodes the video into visual representations, a question encoder ($E_Q$) that encodes the question into text representations, a video-text interaction module ($H$) that captures the cross-modal correlations, and an answer predictor ($F$) that outputs the predictions based on the fused video-text representations.

Taking OEQA as an example, given  a video $V$ and a question $Q$, the video encoder  $E_V$ and the question encoder $E_Q$ extract visual and text representations, respectively, i.e.,
\begin{equation}
    R_V=E_V(V),R_Q=E_Q(Q).
\end{equation}
And then, the video-text interaction module $H$ takes the visual representation $R_V$ and the text representation $R_Q$ as input for modality alignment, and it outputs the fused feature $R$, which is exploited for answer prediction, as follows,
\begin{align}
    &R=H(R_V,R_Q),\\
    &p(a|V,Q)=F(R,R_V,R_Q).
\end{align}
Note that the above modules are just abstractions of specific network structures, which can be instantiated to various models. 
For example, convolutional neural networks (CNN) and the Vision Transformer (ViT) are widely used as the video encoder, while long short-term memory (LSTM) and the Transformer \cite{vaswani2017attention} are commonly employed as the question encoder. Besides, diverse techniques are utilized for video-text interaction, such as bilinear attention network (BAN) \cite{kim2018bilinear} and graph neural networks (GNN). 
Meanwhile, the proposed framework is agnostic to  the VideoQA model and can be used to improve the performance of existing methods as a plug-and-play strategy. In the following section, we introduce our framework without specifying the model structure.

\section{The Proposed Method}

\subsection{\mr{Causal Perspective for VideoQA}}

\begin{figure}[t]
    \centering
    \includegraphics[width=0.75\linewidth]{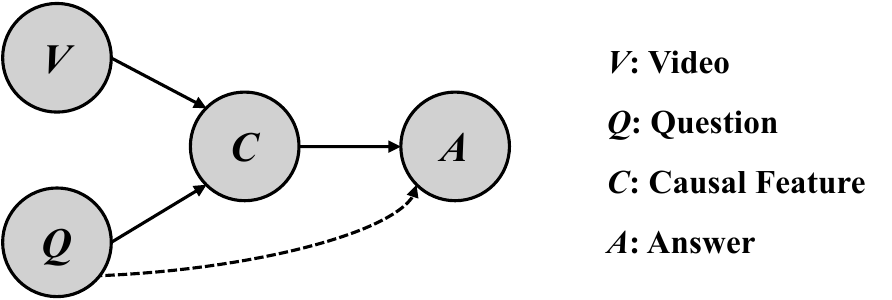}
    \caption{\mr{The casual graph of VideoQA.}}
    \label{graph}
\end{figure}

\mr{Similar to IGV \cite{li2022invariant}, we have indeed carried out causal analysis in the context of VideoQA. To illustrate the relationships among the key variables involved in VideoQA, we have designed a causal graph, which is presented in Fig. \ref{graph}. This causal graph helps to visualize the cause-and-effect connections between four important elements: the input video $V$, the input question $Q$, the causal multimodal feature $C$, and the ground-truth answer $A$. Specifically, a detailed breakdown of the causal relationships depicted in the graph is as follows:}
\begin{itemize}
    \item \mr{$V\rightarrow C \leftarrow Q$: The causal multimodal feature $C$ is determined by the combination of the video $V$ and the question $Q$. In other words, $C$ distills the joint semantic information from both the video and the question. This process is crucial as it extracts the relevant multimodal information needed for answering the question accurately.}
    \item \mr{$C\rightarrow A$: The ground-truth answer $A$ should be inferred from the causal multimodal feature $C$. Since $C$ contains the complete information required to answer the question correctly, the model should rely on $C$ to generate the appropriate answer.}
    \item \mr{$Q \dashrightarrow A$: This represents the spurious causality between the input question $Q$ and the answer $A$. Such spurious correlations often stem from language biases that deep models can easily exploit. For instance, in a question like ``What is the baby doing?", the model might simply guess the answer as ``crying" without actually analyzing the video content. This might be due to the presence of a strong spurious correlation within the dataset, or perhaps the model faces difficulties in accurately identifying the baby within the video.}
\end{itemize}

\mr{In our work, our main objective is to break this spurious correlation between $Q$ and $A$. We achieve this by making interventions on the questions and compelling the model to admit its ignorance when dealing with certain questions. Instead of just minimizing the empirical risk based on the original question-answering pairs, we introduce additional data. This extra data serves to encourage the model to truly utilize the causal multimodal feature $C$ for making accurate answer predictions. By doing so, we aim to improve the model's performance and reduce its reliance on spurious correlations, thereby enhancing the overall quality of the VideoQA system.}

\subsection{Admitting Ignorance for VideoQA}

We argue that the VideoQA model tends to remember the correlations between the question and the answer when the learned video-text alignment is unsatisfactory and untrustworthy. In this work, we aim to break such correlations in the training set by making interventions to the questions and forcing the model to admit its ignorance. By training the model with the intervened questions and forcing it to predict ``unknown'', we can obtain more general correlations between the questions and answers and achieve a more robust alignment between the question and the video.

Specifically, during training, given an OEQA training example $(V,Q,a)$ (or an MCQA example $(V,Q,\mathcal{A},a)$), we make an intervention to the question $Q$ and obtain the intervened question $Q'$. We force the model to predict a different answer instead of the correct answer $a$ based on the intervened input $(V,Q')$ (or $(V,Q',\mathcal{A})$ for MCQA). Nevertheless, this leaves us with two challenges: 1) how to make interventions to the questions, and 2) what the predicted answer should be after intervention. In this work, we address these challenges by proposing a unified methodology for question intervention and answer prediction for different types of VideoQA tasks and VideoQA datasets of diverse characteristics. Regarding question intervention, we present two approaches, global replacement (displacement) and local replacement (perturbation), explained as follows.

\noindent\textbf{Global Replacement (Displacement).} Specifically, the question in a video-question pair is replaced with a question from other video-question pairs. Note that we avoid replacing the question with general ones, such as ``\textit{What is the man doing?}'', to prevent the displaced question from remaining meaningful for the video. \mr{More specifically, in practice, we have set a rule to avoid replacing questions that strictly adhere to the template “What does the [SOMEONE] doing?”. By leveraging the template-based nature of our datasets, we can programmatically scan and detect questions in this format, ensuring that they are not subjected to the displacement strategy\footnote{\mr{The implementation of this exclusion is facilitated by the nature of the datasets we employ. Our datasets are template-based, which means that the structure of each question is highly organized and predictable. This inherent structure allows us to easily identify general questions.}}.}
The aim of global replacement is to compel the model to learn coarse correspondences between the video and the question. In other words, if the model successfully acknowledges its lack of knowledge about the globally-intervened input, it indicates that the model has developed the ability to understand both the question and the video in a coarse manner.

\noindent\textbf{Local Replacement (Perturbation).} This strategy changes only certain crucial words in the questions, which can be regarded as perturbations to questions. In this work, we consider as crucial the words that are important to identify and locate the visual information in both temporal and spatial dimensions related to question answering in the video. For instance, the subject (e.g., ``\textit{man}'', ``\textit{boy}'', ``\textit{dog}''), adjectives (e.g., ``\textit{white}'', ``\textit{big}'', ``\textit{left}''), and prepositions (e.g., ``\textit{above}'', ``\textit{before}'') in the questions are deemed as crucial, as they are necessary to pinpoint the right part of the video to find answers. Note that the perturbed question and the original one are almost the same except for a certain word. In this case, the model is expected to capture the fine-grained correspondence between the video and the question. In other words, the model should have the ability to discover subtle inconsistencies between the two modalities.

Another special consideration is that when the perturbation changes the meaning of the question little, such as ``\textit{child}''$\rightarrow$``\textit{kid}'' and ``\textit{woman}''$\rightarrow$``\textit{lady}'', we regard such perturbation as an augmentation of the question and keep the original answer. To identify such perturbations, we compute the semantic distance between the original question and the perturbed one. We then set a threshold, forcing the model to admit its ignorance to only the questions above it. This type of question perturbation (augmentation) can also be useful as it 1) helps the model distinguish between significant and minor semantic changes and 2) augments the training data for robust question understanding.

\mr{We acknowledge that there is a possibility that the intervened questions may not necessarily lead to an ``unknown" answer with respect to the video content. However, we would like to clarify that such cases are extremely infrequent. In the datasets we have used, when videos contain multiple individuals, questions typically include an adjective to uniquely identify the person in question. For instance, instead of a simple question like ``What is the man doing?", the question might be ``What is the man in red doing?". In such scenarios, when we perform local replacement by simply changing the subject (e.g., from ``man" to ``woman" while keeping the adjective ``red"), the new question will usually become unrelated to the video content, thus requiring an ``unknown" answer. The inherent characteristics of the datasets we employed ensure that the situation almost never occurs.}

In terms of answer designing for intervened video-question pairs, we propose different strategies of admitting ignorance for different types of VideoQA tasks, including multi-choice VideoQA (MCQA) and open-ended VideoQA (OEQA). We elaborate each of them as follows.

\subsection{Admitting Ignorance for MCQA}

As described in Section \ref{vqaf}, MCQA aims to choose the correct answer from given options. Formally, we assume the model is expected to choose an answer from $\mathcal{A}=\{a_i\}_{i=1}^N$ based on the video $V$ and the question $Q$. If the model is not compelled to acknowledge its lack of knowledge, it will select the answer of the highest correctness score from the given options. However, we force the model to predict ``unknown'' when the question $Q$ is intervened. 

To achieve this goal, we manually add the option, ``not given'', into $\mathcal{A}$ for all training examples. And then, for the intervened video-question pairs, we force the model to select the ``not given'' option. Formally, $\mathcal{A}$ is augmented with $a_{N+1}=\mathrm{``not~given\text{''}}$  as follows,
\begin{equation}
    \mathcal{A}'=\mathcal{A}\cup\{a_{N+1}\}=\{a_i\}_{i=1}^{N+1}.
\end{equation}
Then, during training, we change the correct answer to the intervened video-question pairs to $a_{N+1}$, and keep the correct answer to the unchanged pairs. 

Besides, to prevent the scenario where $\mathcal{A}$ still contains the correct answer to the intervened question, we replace the $N-1$ wrong options with respect to the original question with options randomly from all options in the training set. That is to say, if the question is intervened, the model would choose from the option set defined as $\mathcal{A}''=\{b_1,\cdots,b_{N-1},a_{i^*},a_{N+1}\}$, 
% \begin{equation}
%     \mathcal{A}''=\{b_1,\cdots,b_{N-1},a_{i^*},a_{N+1}\},
% \end{equation}
where $\{b_i\}_{i=1}^{N-1}$ are randomly-sampled options, $a_{i^*}$ is the correct answer to the original question, and $a_{N+1}=\mathrm{``not~given\text{''}}$. The model is forced to select $a_{N+1}$. 

\mr{The reason we keep the correct answer of the original question, \(a_{i^*}\), in the new option set \(\mathcal{A}''\) is twofold. First, by including \(a_{i^*}\), we introduce an element of confusion for the model. Since the question has been changed, the model should not be able to rely solely on the presence of the original correct answer to make a decision. Second, we want to force the model to admit its ignorance about the new state of the question. Instead of simply choosing the original correct answer \(a_{i^*}\) out of habit or due to the presence of the familiar answer in the option set, the model is now required to consider the option \(a_{N+1} = \mathrm{``not~given"}\). This way, we can more accurately assess the model's ability to recognize when it doesn't have the necessary information to make a proper choice and avoid false positives that could occur if the model just selects the original correct answer despite the question being modified. This approach helps us better understand the model's true understanding and decision-making process when dealing with questions that have been altered, rather than having the model be influenced by the remnants of the original question's options.}

\mr{Note that we sample answer options from an extensive pool that encompasses the combined options of all questions in our dataset. This vast options pool contains a rich variety of semantic elements, including nouns, verbs, adjectives, and more, which significantly diversifies the nature of the options. Given this high level of semantic diversity, the likelihood that the randomly sampled options would accurately correspond to an intervened (perturbed) question is extremely low. As the options cover a wide range of concepts and entities, it becomes statistically improbable for them to match the specific requirements of a modified question.}

\mr{In our proposed framework, the additional ``not given” option is specifically intended to denote the situation where ``the answer cannot be inferred from the video”. Given that we have engineered the question-video relationships to eliminate valid answers, the model does not need to differentiate between the case of ``the answer does not exist in the option” and ``the answer cannot be inferred from the video”. In essence,  due to our augmentation method, these two seemingly different scenarios converge into a single case that the ``not given” option addresses. }

\begin{figure}[t]
    \centering
    \includegraphics[width=\linewidth]{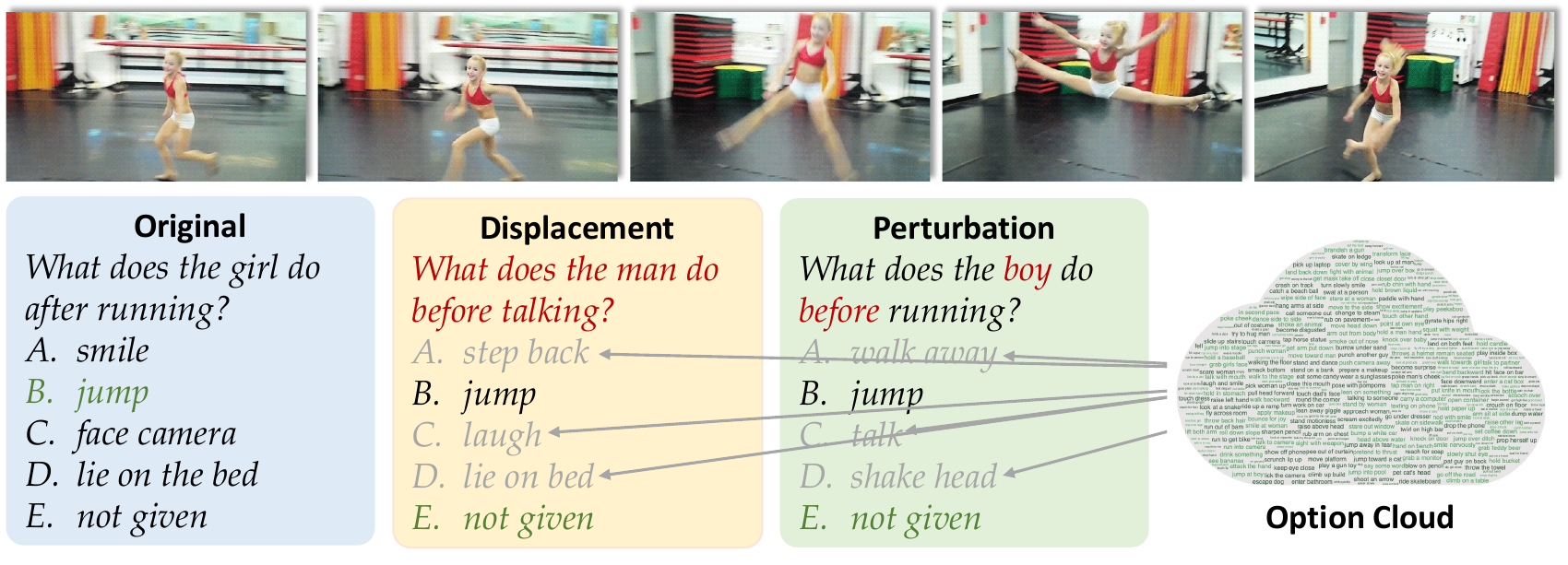}
    \caption{Admitting ignorance for multi-choice VideoQA: a ``not given'' option is added for each question, and other options except the correct one are substituted by random options from the option pool for the intervened questions.}
    \label{exam2}
\end{figure}

To summarize, as shown in Fig. \ref{exam2}, given a training example $(V,Q,\mathcal{A},a_{i^*})$, we augment and intervene it to $(V,Q,\mathcal{A}',a_{i^*})$ and $(V,Q',\mathcal{A}'',a_{N+1})$, respectively. The training follows the approach as described in Section \ref{vqaf}.

\subsection{Admitting Ignorance for OEQA}

A naive strategy for handling ignorance in OEQA is to introduce an ``unknown'' class alongside the original classes. However, this approach can lead to class imbalance issues when a significant number of interventions are applied to the questions. In such cases, the model may become overfitted to the ``unknown'' class, resulting in naive ``unknown'' predictions. To address this problem, we make a modification to the last prediction layer of the VideoQA model. We introduce an additional score that indicates whether the input video-question pair has been subject to an intervention.

Formally, given a $C$-class VideoQA model ($f_{\theta}$), we modify it and make it predict the logits of $C+1$ dimensions ($\boldsymbol{h}$), where the first $C$ dimensions are for answer prediction, and the last dimension indicates whether the model acknowledges its ignorance, i.e.,
\begin{align}
&\boldsymbol{h}=f_{\theta}(V,Q_v)\in\mathbb{R}^{C+1},\\
&p(a|V,Q_v)=\mathrm{softmax}(\boldsymbol{h}_{1:C}),\\
&p(i|V,Q_v)=\mathrm{sigmoid}(\boldsymbol{h}_{C+1}), \label{h-1}
\end{align}
where $Q_v\in\{Q,Q'\}$ is the input question, randomly set to the original question $Q$ or the intervened one $Q'$. During training, we require the model to 1) admit its ignorance when the question is intervened and 2) predict the answer correctly when the question remains unchanged. Specifically, we define the loss as the linear combination of two cross-entropy losses as follows,
\begin{equation}
    \mathcal{L}=-(1-d)\log p_{a^*}-(d\log p_i+(1-d)\log (1-p_i)),\label{loss}
\end{equation}
where $p_{a^*}=p(a=a^*|V,Q)$, $a^*$ is the correct answer, $p_i=p(i|V,Q_v)$, and $d=D(Q,Q_v)\in[0,1]$ is the semantic distance between the original question $Q$ and the input the question $Q_v$ ($D$ is a model that computes semantic distance, elaborated in Section \ref{qp}). The first part of the loss compels the model to predict the correct answer, and it comes into effect when the question is slightly intervened in a semantic sense. Meanwhile, the second part always requires the model to express whether it is ignorant. Note that we do not use hard binary cross-entropy for admitting ignorance; instead, we employ soft labels based on the intervention degree (i.e., the question distance $d$) for better generalization.

\subsection{From Admitting Ignorance to Presenting Answers}
\label{cl}

During training with the proposed strategy, three types of data are involved: the original video-question pairs, the pairs with locally-replaced questions, and the pairs with globally-replaced questions. Intuitively, the difficulty of addressing these questions decreases from the first to the last. 
Specifically, for the pairs with globally-replaced questions, the model is required to merely indicate whether it is ignorant, and it is easy because all the model needs to do is find the coarse-grained semantic inconsistency between the video and the question. Regarding the pairs with locally-replaced questions, they are a bit harder because the model is required to perceive subtle semantic differences between the video and the question. In contrast to the intervened video-question pairs, the original pairs are the most challenging because the model is expected to provide specific answers instead of simply indicating its ignorance.
Considering this property, we propose an \textbf{ignorance-diminishing curriculum learning framework}. Specifically, during training, we set a probability for each epoch, $p(e)$, which represents the probability of modifying the questions (either through global replacement or local replacement) in the video-question pairs. We then randomly replace the questions with this probability. As training progresses, we gradually decrease $p(e)$ to 0, which means that no questions will be changed by the end of training. With this strategy, the model has more opportunities to simply admit its ignorance and learn from easy data in the early stages. In the later phases, it is required to predict answers more frequently and learn from challenging data.

\begin{algorithm}[tbp]
\caption{Admitting-ignorance (AI) training framework for open-ended VideoQA.}\label{alg}
\KwData{Training set $\left\lbrace (V_n,Q_n,a_n)\right\rbrace_{n=1}^N$, VideoQA model $f_{\theta}$, semantic distance model $D$, epoch $E$, scheduler $p(e)$, and learning rate $\gamma$.}
\For{$e=1$ to $E$}{
\While{not done}{
Sample an example $(V,Q,a)$\\
\eIf{$\mathrm{RAND}(0,1)<p(e)$}
    {
    \tcc{Intervene in Questions}
    $Q'=\mathrm{INTERV}(Q)$\\
    \tcc{Semantic Distance Between Quesitons}
    $d=D(Q,Q')$ \\
    $Q=Q'$
    }
    {
    $d=0$\\
    }
\tcc{Prediction and Admit Ignorance}
$p_{a},p_i=f_{\theta}(V,Q)$ \\
\tcc{Loss Function}
$\mathcal{L}=-(1-d)\log p_{a}
-(d\log p_i+(1-d)\log (1-p_i))$ \\
\tcc{Optimization}
$\theta=\theta-\gamma\nabla_{\theta}\mathcal{L}$
}}
\end{algorithm}

Furthermore, \textit{our prior experiments show that, when using a fixed probability for replacement during training, the model tends to naively admit its ignorance regarding the challenging video-question pairs in the testing set, resulting in poor accuracy}. We assume that the reason is a fixed probability leading to underfitting of the original data. In this case, our ignorance-diminishing curriculum learning framework is necessary for better performance. Another strategy involves learning from the original and intervened pairs simultaneously \cite{li2022invariant,li2023transformer}, but such an approach requires much more training time and GPU memory. In contrast, our curriculum learning achieves a satisfactory balance between computational cost and performance. The pseudo code for our framework is shown in Algorithm \ref{alg}\footnote{We use SGD as an example. Other optimizers can also be applied.}.

\subsection{Apply to VQA Models}

As we elaborate on the proposed framework, our focus lies on data augmentation/intervention and loss design. We do not specify the VideoQA model structure, which means our method is model-agnostic. Most of the current popular deep neural networks for VideoQA can be seamlessly integrated into our framework, including models explicitly designed for VideoQA \cite{jang2017tgif,seo2021attend,xiao2022video,buch2022revisiting} and multi-modal foundation models pretrained on large-scale data \cite{zellers2021merlot,fu2021violet,zeng2022x,wang2022internvideo,wang2023all}. In this work, without loss of generality, we integrate InternVideo \cite{wang2022internvideo}, a pretrained video-text foundation model, into our framework. We choose InternVideo for two primary reasons: 1) it is a typical model comprising general modules for VideoQA, and 2) it is one of the state-of-the-art pretrained models for various downstream tasks, including video-question answering, video retrieval, and visual language navigation. By integrating this powerful model into our framework, we can verify that our method can further improve performance, and the improvement brought by it is orthogonal to that resulting from pretraining.

Specifically, InternVideo comprises a video encoder, a text encoder (based on CLIP pretraining \cite{radford2021learning}), a multimodal alignment module for video-text fusion, and a prediction head. It leverages a substantial amount of unsupervised and supervised data for pretraining, including action recognition \cite{carreira2017quo}, video captioning \cite{miech2019howto100m}, action localization \cite{gu2018ava}, and visual retrieval \cite{bain2021frozen}. To fine-tune InternVideo with the proposed framework, we retain the main modules, and for different types of VideoQA tasks, we modify only the prediction head and the input video-question pairs, as described in the previous sections. We refer to InternVideo fine-tuned with our framework as ``\textbf{AIQA}'', distinguishing it from the counterpart fine-tuned in the straightforward manner. It is important to note that we also compare other models trained with and without our framework to demonstrate the impact of our contribution in Section \ref{gen}.

\noindent\textbf{Training and Inference Complexity.} We would like to emphasize that, during training, we replace a proportion (which decreases as training proceeds) of the original questions with the intervened ones instead of adding them to the dataset. Therefore, the training time of our framework remains the same compared to naive training. Meanwhile, since we only modify the training data and keep the testing data and model structure unchanged, the inference time on the testing set also remains the same.

\noindent\textbf{Generalization to ImageQA.} As we focus solely on text-side debiasing, our method is visual-form-agnostic and can easily be generalized to ImageQA, which shares the same goal and formulation as VideoQA, except that it finds answers from images instead of videos. 

\section{Experiments}

In this section, we conduct experiments to show the effectiveness of our method. Specifically, we first explain the experiment settings. We then make comparisons between our method and the state of the art. Furthermore, we apply our method to other models to show its generalization ability. Finally, we conduct more analysis regarding the ability of admitting ignorance and hyperparameters.

\subsection{Experiment Settings}

\subsubsection{Datasets and Question Perturbation.}
\label{qp}

Two types of datasets are utilized for the evaluation: the multi-choice datasets, including TGIF-\textit{Action} \cite{jang2017tgif}, TGIF-\textit{Transition} \cite{jang2017tgif}, and NExT-QA \cite{xiao2021next}, and the open-ended ones, including TGIF-\textit{FrameQA} \cite{jang2017tgif}, MSVD-QA \cite{xu2017video}, and MSRTT-QA \cite{xu2017video}. 

Besides the task formulation, the questions from different datasets take different forms. Specifically, questions from TGIF-\textit{Action} and TGIF-\textit{Transition} are in fixed forms, generated using fixed templates such as ``\textit{What does SOMEONE do SOME-NUMBER times?}'' and ``\textit{What does SOMEONE do before/after SOME-ACTION?}''\footnote{``\textit{SOMEONE}'' and ``\textit{SOME-ACTION}'' could represent short phrases such as ``\textit{the girl in red}'' and ``\textit{closing eyes}'', respectively.}  For these types of questions, the perturbation is implemented by manually replacing the crucial words (including the subjects, modifiers of the subjects, ``\textit{SOME-NUMBER}'', and ``\textit{before/after}'') with other frequent words in the datasets, such as ``\textit{boy}''$\rightarrow$``\textit{woman}'', ``\textit{red}''$\rightarrow$``\textit{black}'', and ``\textit{2 times}''$\rightarrow$``\textit{5 times}''. Note that such manual replacement is possible because the question structures are fixed and evident, making it straightforward to identify the crucial parts.

\begin{table*}[t]
\caption{The comparisons (Accuracy, \%) with the state of the art, including the multi-choice VideoQA and the open-ended ones. The compared methods include the conventional VideoQA models and the large pretrained (PT) video-text models fine-tuned on VideoQA datasets. $^*$ means the result is re-implementation.} 
\label{comp}
\center
% \vspace{-1em}
% \resizebox{\textwidth}{!}{
\begin{tabular}{@{}lcllllll@{}}
\toprule
\multirow{2}{*}{Method}                & \multirow{2}{*}{PT} & \multicolumn{3}{c}{Multi-Choice}                          & \multicolumn{3}{c}{Open-Ended}              \\ \cmidrule(l){3-5} \cmidrule(l){6-8} 
                                       &                     & TGIF-\textit{Action}& TGIF-\textit{Transition}& NExT-QA & TGIF-\textit{FrameQA}& MSVD-QA& MSRVTT-QA\\ \midrule
MASN \cite{seo2021attend}              &                     & 84.4                 & 87.4                     & 52.2    & 59.5                  & 38.0    & 35.2      \\
HQGA \cite{xiao2022video}              &                     & 76.9                 & 84.6$^*$& 51.8    & 57.5$^*$& 39.7$^*$& 38.6      \\
B2A \cite{park2021bridge}              &                     & 75.9                 & 82.6                     & ---     & 57.5                  & 37.2    & 36.9      \\
IGV \cite{li2022invariant}             &                     & 78.5                 & 85.7                     & 51.3    & 52.8                  & 40.8    & 38.3      \\
HOSTER \cite{dang2021hierarchical}     &                     & 75.6                 & 82.1                     & ---     & 58.2                  & 39.4    & 35.9      \\ \midrule
ClipBERT \cite{lei2021less}            & \cmark          & 82.8                 & 87.8                     & ---     & 60.3                  & ---     & 37.4      \\
VIOLET \cite{fu2021violet}             & \cmark          & 92.5                 & 95.7                     & ---     & 68.9                  & 47.9    & 43.9      \\
All-in-one \cite{wang2023all}          & \cmark          & 94.3$^*$& 96.6$^*$& ---     & 64.2                  & 46.5    & 42.9      \\
InternVideo$^*$ \cite{wang2022internvideo}& \cmark          & \underline{95.2}& \underline{97.1}& \underline{54.6}& \underline{71.8}& \underline{55.5}    & \underline{46.4}\\ \midrule
AIQA (Ours)                                   & \cmark          & \textbf{97.1} {\color[HTML]{036400}(+1.9)}& \textbf{98.8} {\color[HTML]{036400}(+1.7)}& \textbf{56.5}  {\color[HTML]{036400}(+1.9)}& \textbf{73.1} {\color[HTML]{036400}(+1.3)}& \textbf{56.7} {\color[HTML]{036400}(+1.2)}&    \textbf{47.5} {\color[HTML]{036400}(+1.1)}\\ \bottomrule
\end{tabular}
% }
\end{table*}

\begin{table*}[t]
\caption{The comparisons (\%) between the models fine-tuned (trained) with and without admitting-ignorance (AI). The improvement ($\Delta$) is also provided.} 
\label{ai}
\center
% \vspace{-1em}
\begin{tabular}{@{}lcccccccccccc@{}}
\toprule
\multirow{2}{*}{Dataset} & \multicolumn{6}{c}{Multi-Choice}                                                        & \multicolumn{6}{c}{Open-Ended}                                                          \\\cmidrule(l){2-7}\cmidrule(l){8-13}
                        & \multicolumn{3}{c}{TGIF-\textit{Action}}        & \multicolumn{3}{c}{TGIF-\textit{Transition}}        & \multicolumn{3}{c}{TGIF-\textit{FrameQA}}        & \multicolumn{3}{c}{MSVD-QA}                   \\
                        % \cmidrule(l){2-4}\cmidrule(l){5-7}\cmidrule(l){8-10}\cmidrule(l){11-13}
                        \midrule
        AI                & \xmark    & \cmark   & $\Delta$                     & \xmark    & \cmark   & $\Delta$                     & \xmark    & \cmark  & $\Delta$                     & \xmark    & \cmark   & $\Delta$                     \\ \midrule
HQGA                    & 76.9 & 78.0 & {\color[HTML]{036400}+1.1} & 84.6 & 85.3 & {\color[HTML]{036400}+0.7} & 57.5 & 58.2 & {\color[HTML]{036400}+0.7} & 39.7 & 41.2 & {\color[HTML]{036400}+1.4} \\
All-in-one-T            & 90.1 & 91.8 & {\color[HTML]{036400}+1.7} & 95.5 & 96.8 & {\color[HTML]{036400}+1.3} & 53.9 & 55.1 & {\color[HTML]{036400}+1.2} & 32.1 & 33.2 & {\color[HTML]{036400}+1.1} \\
All-in-one-S            & 93.4 & 95.0 & {\color[HTML]{036400}+1.6} & 96.1 & 97.2 & {\color[HTML]{036400}+1.1} & 62.5 & 63.3 & {\color[HTML]{036400}+0.8} & 41.7 & 42.6 & {\color[HTML]{036400}+0.9} \\
All-in-one-B            & 94.3 & 95.3 & {\color[HTML]{036400}+1.0} & 96.1 & 97.2 & {\color[HTML]{036400}+1.1} & 64.2 & 66.4 & {\color[HTML]{036400}+2.2} & 46.5 & 47.8 & {\color[HTML]{036400}+1.3} \\
InternVideo-B           & 92.9 & 95.3 & {\color[HTML]{036400}+2.4} & 97.0 & 98.4 & {\color[HTML]{036400}+1.4} & 67.4 & 68.2 & {\color[HTML]{036400}+0.8} & 51.1 & 52.7 & {\color[HTML]{036400}+1.6} \\
InternVideo-L           & 95.2 & 97.1 & {\color[HTML]{036400}+1.9} & 97.1 & 98.8 & {\color[HTML]{036400}+1.7} & 71.8 & 73.1 & {\color[HTML]{036400}+1.3} & 55.5 & 56.7 & {\color[HTML]{036400}+1.2} \\ \bottomrule
\end{tabular}
\end{table*}

For the free-form questions, it is intractable to analyze their structures and lexical components manually. Fortunately, with the advancements in large language models, automatic text perturbation becomes possible \cite{feder2022causal,calderon2022docogen,wu2021polyjuice}. In this paper, we utilize Polyjuice \cite{wu2021polyjuice}, a model fine-tuned based on GPT-2 \cite{radford2019language}, to generate perturbations for free-form questions. In Polyjuice, various control codes are designed to guide the generation, including negation, resemantic, etc. Considering both the quality (fluency and diversity) of the generated text and our purpose, we choose the following control codes: 1) lexical, which involves modifying one word without changing the Part-of-Speech tags; 2) shuffle, which entails moving/swapping key entities around the sentence; and 3) quantifier, which involves modifying the number in the sentence. For each pair of the original question $Q$ and the perturbed one $Q'$, we use Sentence-BERT \cite{reimers-2019-sentence-bert} to calculate their semantic distance $d=D(Q,Q')$.

\subsubsection{Training Details.}

We follow the suggested fine-tuning settings of InternVideo, including the learning rate, number of epochs, and batch size. Regarding the proposed training framework, several settings and hyper-parameters are crucial to the final performance. Specifically, we set $p(e)$ in Section \ref{cl} to a quadratically-decreasing function as follows,
\begin{equation}
p(e)=\frac{p_r}{E^2}\left(e-E\right)^2,e\in[1..E],
\end{equation}
where $E$ is the number of epochs, and $p_r$ is the initial probability of replacing the question. With this design, $p(e)$ decreases from $p_r$ to 0 in a quadratic manner, allowing the model to start with easier tasks and gradually shift its focus towards the original task. The quadratic design is chosen to ensure that the ``unknown'' prediction does not dominate the answer distribution in the later training phases; otherwise, the accuracy of the original data could be compromised. We set $p_r$ to various values on different datasets based on validation accuracy. Another important hyper-parameter is the ratio of displacement/perturbation in all question replacements, also determined through validation. All experiments are conducted using PyTorch on NVIDIA A100 GPUs. All settings remain consistent, whether with or without our method, to guarantee fair comparisons.

\subsection{Comparisons to Existing Methods}

We compare our method with existing models, including conventional VideoQA models (without pretraining) such as MASN \cite{seo2021attend} and HQGA \cite{xiao2022video}, as well as pretrained foundation models that are fine-tuned on VideoQA, such as All-in-one \cite{wang2023all} and InternVideo \cite{wang2022internvideo}, some of which represent the state of the art in VideoQA tasks. The comprehensive comparisons are illustrated in TABLE \ref{comp}.

As shown in TABLE \ref{comp}, pretrained models outperform conventional methods to a large extent, especially on the sub-tasks of TGIF and MSVD. Despite the significant performance achieved by pretrained models, our method further notably enhances accuracy. Specifically, the improvements with respect to InternVideo on all datasets are more than 1\%, with the most noticeable improvements observed on multi-choice datasets. We assume that the reason for the better results on these datasets is the presence of strong but spurious correlations between questions and answers in the training sets, and our method has a greater impact on breaking such correlations, enabling the model to learn more robust video-text representations for answer prediction. On the other hand, the relatively smaller improvements on other datasets could be attributed to the large size of these datasets, which weakens the spurious question-answer correlations.

\mr{Although our proposed methods have achieved improvements across multiple types of datasets, the enhancements on open-ended questions appear to be relatively modest. There are likely two main reasons for this.
Firstly, the spurious correlations between questions and answers in the training sets are rather weak. As our methods are designed in part to break these correlations, the limited strength of such correlations reduces the effectiveness of our approach in enhancing performance on open-ended questions.
Secondly, the task of modeling ``admitting ignorance" in our framework essentially boils down to an open-set classification problem, which remains a highly challenging and unsolved issue in the field. Currently, our approach of simply adding an additional ``unknown" class to the answer set is a rather simplistic solution. We are confident that by employing more sophisticated techniques for open-set classification, we can achieve more significant improvements. This will be a key area of focus in our future research endeavors. }

\begin{table}[t]
\caption{The ability of models in admitting ignorance to intervened questions, including displacement (\textit{D}) and perturbation (\textit{P}). The evaluation is accuracy (\%) of successfully admitting ignorance.} 
\label{aiacc}
% \vspace{-1em}
\center
% \resizebox{0.4\textwidth}{!}{
\begin{tabular}{@{}lcccc@{}}
\toprule
\multirow{2}{*}{Model} & \multicolumn{2}{c}{TGIF-\textit{Action}} & TGIF-\textit{FrameQA} & MSVD-QA           \\\cmidrule(l){2-3}\cmidrule(l){4-4}\cmidrule(l){5-5}
                       & \textit{D}   & \textit{P}   & \textit{D}  & \textit{D} \\ \midrule
All-in-one-S           & 83.5             & 15.9             & 71.5            & 87.6      \\
All-in-one-B           & 89.7             & 23.6             & 73.2            & 89.3           \\
InternVideo-B          & 40.5             & 39.5             & 77.1            & 94.8           \\
InternVideo-L          & 50.0             & 49.6             & 83.8            & 95.5           \\ \bottomrule
\end{tabular}
% }
\end{table}

\subsection{Generalize to Other Models}
\label{gen}

We have also applied our framework to the conventional VideoQA model, HQGA, as well as pretrained models of various versions, including All-in-one (Tiny, Small, and Base) and InternVideo (Base and Large). The results are shown in TABLE \ref{ai}. As we can see from the table, our method consistently enhances the performance of different models (across various versions), with most of the improvements exceeding 1\%. Furthermore, it is observed that the improvements on the multi-choice datasets are generally greater than those on the open-ended ones. Notably, there are significant improvements for InternVideo-B on TGIF-\textit{Action} and All-in-one-B on TGIF-\textit{FrameQA}.

\subsection{More Analysis}

\noindent\textbf{Is the Model Admitting Ignorance?}
To assess whether the models trained with our framework are capable of acknowledging their ignorance when presented with intervened questions, we apply interventions to the questions in the testing set, similar to the training phase, and evaluate the predictions. Specifically, for multi-choice VideoQA, we expect the models to choose ``not given''. For open-ended VideoQA, we expect the models to exhibit a high activation level\footnote{We set a threshold for the activations, considering those greater than it as successfully acknowledging ignorance.} in the last dimension of the predicted logits (Eq. \ref{h-1}). We demonstrate the ability of two models of two different versions in TABLE \ref{aiacc}. Interestingly, as the results show, even though we apply the ignorance-diminishing curriculum learning strategy, the models can still identify inconsistencies between the question and the video and acknowledge their ignorance. Meanwhile, it is anticipated that perturbations are more challenging to detect than displacements for the models, as the semantic changes from displacements are more significant. 
Furthermore, we remove interventions and keep the ``Unknown'' option in testing. We observe rare selections (0.6\%/0.4\% on TGIF-\textit{Action}/TGIF-\textit{Transition}) of ``Unknown'', which means the model is not biased by ``Unknown'' in training.

\begin{table}[t]
\caption{Comparisons (\%) of different text augmentation methods.} 
\label{aug}
% \vspace{-1.5em}
\center
% \resizebox{0.35\textwidth}{!}{
\begin{tabular}{@{}lll@{}}
\toprule
Augmentation & TGIF-\textit{Action} & TGIF-\textit{FrameQA} \\ \midrule
Baseline     &     95.2            &       71.8          \\ \midrule
Random Drop   &         94.8  {\color[HTML]{CB0000}($-$0.4)}     &      71.2   {\color[HTML]{CB0000}($-$0.6)}    \\
Random Switch  &        94.3  {\color[HTML]{CB0000}($-$0.9)}     &      70.8   {\color[HTML]{CB0000}($-$1.0)}      \\ \midrule
Ours         &     97.1  {\color[HTML]{036400}(+1.9)}        &      73.1   {\color[HTML]{036400}(+1.3)}        \\ \bottomrule
\end{tabular}
% }
\end{table}

\noindent\textbf{Is Naive Text Augmentation Effective?}
Our method can also be considered as text augmentation, wherein we augment the questions in the training set with sophisticatedly designed interventions. To validate the impact of our method, we compare it with other naive text augmentation strategies, such as randomly dropping/switching words in questions. The results of different augmentation methods are reported in TABLE \ref{aug}. As the results show, naive text augmentation strategies negatively affect performance, which, we assume, may be attributed to the introduced uncertainty/ambiguity. In contrast, our method significantly improves accuracy, validating its advantage over naive text augmentation.

 \begin{figure}[tbp]
     \centering
     \begin{subfigure}[b]{0.49\columnwidth}
         \centering
         \includegraphics[width=\columnwidth]{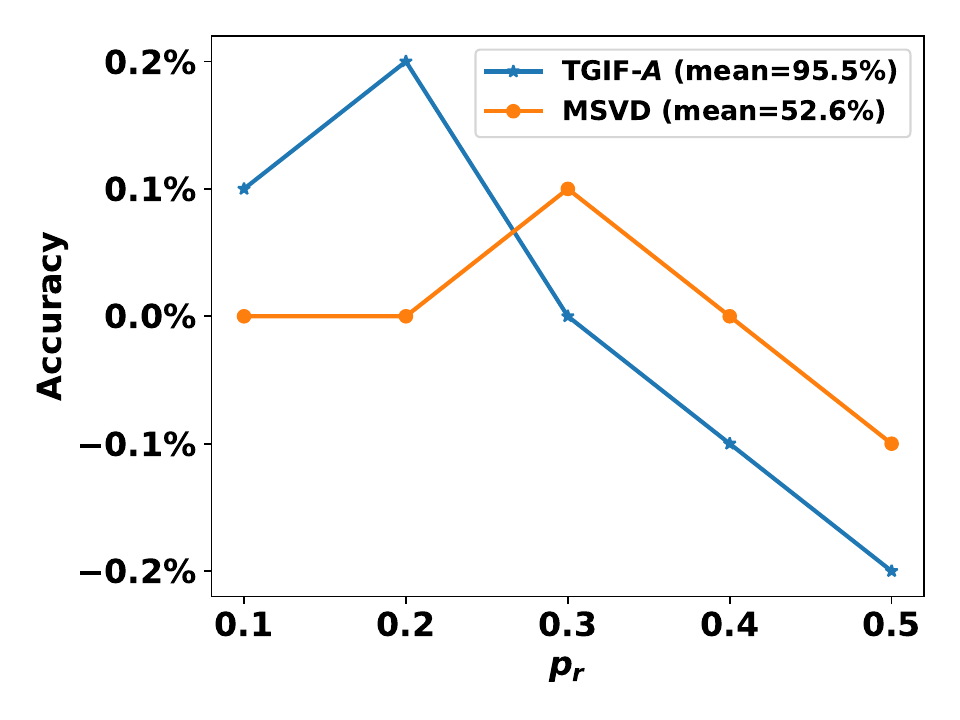}
         \caption{Initial intervention probability.}
         \label{pred}
     \end{subfigure}
     \begin{subfigure}[b]{0.49\columnwidth}
         \centering
         \includegraphics[width=\columnwidth]{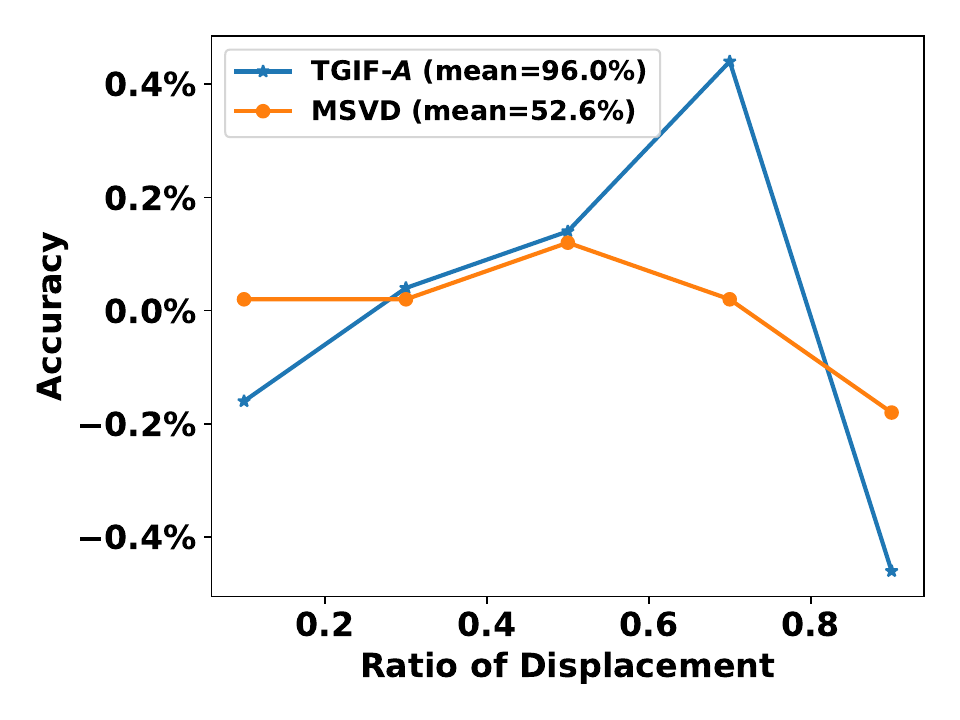}
         \caption{Ratio of displacement.}
         \label{feat}
     \end{subfigure}
        \caption{The impact of the initial intervention probability $p_r$ and the ratio of displacement. Note that the curves illustrate the deviations from the mean accuracy.}
        \label{vis}
% \vspace{-1.5em}
\end{figure}

\noindent\textbf{What is the Impact of the Hyperparameters?}
We also demonstrate the impact of two crucial hyperparameters in our method: the initial intervention probability ($p_r$) and the ratio of displacement to all interventions. We use InternVideo-B for this study. The accuracy on TGIF-\textit{Action} and MSVD with respect to various values of $p_r$ is presented in Fig. \ref{pred}, while keeping the ratio of displacement fixed at 0.5. It has been observed that a larger $p_r$ ($p_r > 0.5$) harms performance, and we assume this is due to the model becoming excessively biased towards the ``unknown'' answer.
Regarding the ratios of displacement, the results are shown in Fig. \ref{feat}, with $p_r$ set to 0.3. The results reveal that for TGIF-\textit{Action}, the highest accuracy is achieved at 0.7, indicating that, for this dataset, displacement plays a more critical role than perturbation. For MSVD, the model performs best when displacement and perturbation are balanced. \mr{When considering only displacement as an intervention, the overall accuracy is slightly lower than when only perturbation is applied. We hypothesize that this is because perturbation compels the model to learn the fine-grained alignment between videos and questions, which is more vital for this task than the coarse-grained alignment that displacement primarily addresses.
Furthermore, as depicted in Fig. 3, the proposed framework exhibits remarkable robustness to the inclusion of hyperparameters. Specifically, it shows that for the initial intervention probability and the ratio of displacement/perturbation, the accuracy undergoes only minimal fluctuations. When we assign appropriate values to these two hyperparameters, the accuracy changes are confined to a narrow range, typically between 0.4\% and 0.8\%. In the context of various datasets, setting the initial intervention probability at 0.3 and establishing the ratio of displacement to perturbation as 1:1 serves as a promising starting point for achieving satisfactory performance. 
}

\begin{table}[t]
\caption{\mr{The comparisons (\%) between the models trained with different intervention probabilities (fixed and our dynamic schedule). The improvement with respective to the original model is also provided.}}
\label{1.2.1}
\center
\begin{tabular}{@{}ccccccccc@{}}
\toprule
\multirow{2}{*}{Prob.} & \multicolumn{4}{c}{Multi-Choice}                                                        & \multicolumn{4}{c}{Open-Ended}                                                          \\\cmidrule(l){2-5} \cmidrule(l){6-9}
                        & \multicolumn{2}{c}{TGIF-\textit{A}}        & \multicolumn{2}{c}{TGIF-\textit{T}}        & \multicolumn{2}{c}{TGIF-\textit{FrameQA}}        & \multicolumn{2}{c}{MSVD-QA}                   \\
                        \midrule
25\%           & 94.1 & {\color[HTML]{036400}+1.2} &  97.4 & {\color[HTML]{036400}+0.4} &  67.7 & {\color[HTML]{036400}+0.2} &  52.1 & {\color[HTML]{036400}+1.0} \\
50\%            & 93.3 & {\color[HTML]{036400}+0.1} &  97.4 & {\color[HTML]{036400}+0.4} &  67.5 & {\color[HTML]{036400}+0.1} &  51.2 & {\color[HTML]{036400}+0.1} \\
75\%            & 92.2 & {\color[HTML]{CB0000}-0.7} &  96.1 & {\color[HTML]{CB0000}-0.9} &  66.2 & {\color[HTML]{CB0000}-1.2} &  50.4 & {\color[HTML]{CB0000}-0.7} \\\midrule
Dynamic           & 95.3 & {\color[HTML]{036400}+2.4} &  98.4 & {\color[HTML]{036400}+1.4} &  68.2 & {\color[HTML]{036400}+0.8} &  52.7 & {\color[HTML]{036400}+1.6} \\
 \bottomrule
\end{tabular}
\end{table}

\begin{table}[t]
\caption{\mr{The comparisons (\%) between the models trained with different schedules for intervention probability. The improvement with respective to the original model is also provided.}}
\label{1.2.2}
\center
\resizebox{\columnwidth}{!}{\begin{tabular}{@{}lcccccccc@{}}
\toprule
\multirow{2}{*}{Sched.} & \multicolumn{4}{c}{Multi-Choice}                                                        & \multicolumn{4}{c}{Open-Ended}                                                          \\\cmidrule(l){2-5}\cmidrule(l){6-9}
                        & \multicolumn{2}{c}{TGIF-\textit{A}}        & \multicolumn{2}{c}{TGIF-\textit{T}}        & \multicolumn{2}{c}{TGIF-\textit{FrameQA}}        & \multicolumn{2}{c}{MSVD-QA}                   \\
                        \midrule
Linear         & 94.0 & {\color[HTML]{036400}+1.1} &  97.5 & {\color[HTML]{036400}+0.5} &  67.7 & {\color[HTML]{036400}+0.3} &  51.6 & {\color[HTML]{036400}+0.5} \\
Exponential           & 95.1 & {\color[HTML]{036400}+2.2} &  98.7 & {\color[HTML]{036400}+1.7} &  68.4 & {\color[HTML]{036400}+1.0} &  52.4 & {\color[HTML]{036400}+1.3} \\
Quadratic           & 95.3 & {\color[HTML]{036400}+2.4} &  98.4 & {\color[HTML]{036400}+1.4} &  68.2 & {\color[HTML]{036400}+0.8} &  52.7 & {\color[HTML]{036400}+1.6} \\
 \bottomrule
\end{tabular}}
\end{table}

\noindent\mr{\textbf{What is the Impact of Curriculum Learning?} We carried out comprehensive experiments to address the lack of a detailed ablation study on intervention probabilities.
We conducted two sets of experiments. The first set focused on fixed intervention probabilities, specifically testing values of 25\%, 50\%, and 75\%. The second set explored alternative schedules, including linear and exponential decay.
The results of the fixed-probability experiments are reported in Table \ref{1.2.1}, while the findings from the alternative schedule experiments are shown in Table \ref{1.2.2}.
The results indicate that using a fixed intervention probability yields lower accuracy compared to our dynamic strategy. Moreover, a high intervention proportion of 75\% negatively impacts the performance.
Among the alternative schedules, the linear decay schedule marginally improves the results. The exponential decay schedule achieves accuracy comparable to that of our proposed strategy.
These experiments validate the necessity and effectiveness of our current strategy.}

\begin{table}[t]
    \caption{\mr{Results on ImageQA datasets of models with and without our training framework.}}
    \label{tab1.1}
\center
\resizebox{\columnwidth}{!}{\begin{tabular}{lcccccc}
\toprule
\multirow{2}{*}{Model} & \multicolumn{3}{c}{VQA V1}    & \multicolumn{3}{c}{VQA V2}    \\\cmidrule(l){2-4} \cmidrule(l){5-7}
                       & Yes/No & Number & Other & Yes/No & Number & Other \\ \midrule
SAN \cite{yang2016stacked}                    & 78.54  & 33.46  & 44.51 & 68.89  & 34.55  & 43.80 \\
SAN+AIQA               &  79.32      &  34.20      &  45.13     &   69.55     &  45.44      &    44.33   \\ \midrule
        MCB \cite{fukui2016multimodal}              &            81.62  &    34.56     &   52.16     &     77.91    &   37.47 &51.76    \\ 
       MCB+AIQA                 &   82.61     &   35.65     &   53.22    &    78.99    &     38.78   &     52.45  \\ \bottomrule
\end{tabular}
}
\end{table}

\noindent\mr{\textbf{What is the Impact on ImageQA?} We conducted supplementary experiments on ImageQA tasks to validate the generalization of our method. We utilized two well-known ImageQA datasets, VQA V1 \cite{antol2015vqa} and VQA V2 \cite{goyal2017making}. To thoroughly evaluate our training framework, we adapted two established methods, SAN \cite{yang2016stacked} and MCB \cite{fukui2016multimodal}, integrating them with our proposed approach. The experimental results are presented in Table \ref{tab1.1}. A clear trend emerges from the table: models trained using our framework consistently outperform their original counterparts. This consistent improvement across different datasets and adapted methods strongly attests to the effectiveness of our approach on ImageQA tasks.}

\noindent\mr{\textbf{What is the Impact of Difference Ways to Admit Ignorance?}  As proposed in existing works, there are different methods for compelling models to ``admit ignorance."
For instance, IGV \cite{li2022invariant} and TIGV  \cite{li2023transformer} operate by compelling the predicted distribution to be uniform. On the other hand, EIGV \cite{li2022equivariant} uses contrastive learning to enforce that the representation of videos that do not support the answer is distinctly different from the representation of useful video-question pairs. We conducted an evaluation to assess the effectiveness of our proposed method against existing approaches.
To carry out this comparison, we integrated our HQGA into the frameworks of IGV, EIGV, and TIGV. These frameworks represent different strategies for prompting models to acknowledge their lack of knowledge. We then evaluated the performance of these integrated models across multiple datasets, with the results summarized in Table \ref{4.3.2}.
The experimental results reveal several key findings: 1) Our method demonstrates performance that is on par with EIGV and TIGV. This shows that our approach can achieve competitive results in enabling models to admit ignorance, matching the capabilities of these well-regarded existing methods. 2) Across all the evaluated datasets, our method consistently outperforms IGV. This indicates that our approach provides more effective training signals for the model to recognize and admit its lack of knowledge. In addition, a significant advantage of our method lies in its implementation. Unlike the compared methods (IGV, EIGV, and TIGV), which necessitate the addition of extra modules and multiple branches of video feature extraction, our method simply augments the training data. This minimal modification to the model architecture makes our method more efficient and straightforward to implement.}

\begin{table}[t]
\caption{\mr{The comparisons (\%) between the models trained with different ways to admit ignorance.}} 
\label{4.3.2}
\center
\begin{tabular}{@{}lcccc@{}}
\toprule
\multirow{2}{*}{Method} & \multicolumn{2}{c}{Multi-Choice}  & \multicolumn{2}{c}{Open-Ended}  \\ \cmidrule(l){2-3}  \cmidrule(l){4-5} 
                        & TGIF-\textit{A} & TGIF-\textit{T} & TGIF-\textit{FrameQA} & MSVD-QA \\ \midrule
IGV    \cite{li2022invariant}                 & 77.5            & 84.6            & 57.7                  & 40.5    \\
EIGV  \cite{li2022equivariant}                  & 77.9            & 85.2            & 58.0                  & 41.4    \\
TIGV   \cite{li2023transformer}                 & 78.1            & 85.1            & 58.4                  & 40.6    \\
AIQA    (Ours)                & 78.0            & 85.3            & 58.2                  & 41.2    \\ \bottomrule
\end{tabular}
\end{table}

\begin{table}[t]
\caption{The comparisons (\%) between VideoLLaMA2 (7B) and the finetuned model with our framework. The improvement is also provided.} 
\label{3.4.1}
\center
\begin{tabular}{lll}
\toprule
\multirow{2}{*}{Model} & Multi-Choice     &  Open-Ended                                                                                                                                                                                                                     \\ \cmidrule(l){2-2} \cmidrule(l){3-3}
                        & MV-Bench \cite{li2024mvbench}                                              & MSVD-QA \cite{xu2017video}                                                                                                                   \\ \midrule
 VideoLLaMA2 \cite{cheng2024videollama}                               &  53.4   &71.7 \\ 
VideoLLaMA2+AIQA                                    & 53.9 {\color[HTML]{036400}(+0.5)}& 72.3 {\color[HTML]{036400}(+0.6)}\\ 
\bottomrule
\end{tabular}
\end{table}

\begin{table}[t]
\caption{\mr{The comparisons (\%) between different answer set design. AIQA$^*$ means that the answer set does not contain the original correct answer. The improvement with respective to the original model is also provided.} }
\label{1.3.1}
\center
\begin{tabular}{llll}
\toprule
\multirow{2}{*}{Method} & \multicolumn{3}{c}{Multi-Choice}                                                                                                                                                                                                                             \\ \cmidrule(l){2-4}
                        & TGIF-\textit{Action}                                              & TGIF-\textit{Transition}                                          & NExT-QA                                                                            \\ \midrule
AIQA                                       & 97.1 {\color[HTML]{036400}(+1.9)}& 98.8 {\color[HTML]{036400}(+1.7)}& 56.5  {\color[HTML]{036400}(+1.9)}\\ 
AIQA$^*$                                       & 96.7 {\color[HTML]{036400}(+1.5)}& 98.5 {\color[HTML]{036400}(+1.4)}& 56.3  {\color[HTML]{036400}(+1.7)}\\ 

\bottomrule
\end{tabular}
\end{table}

\noindent\mr{\textbf{What is the Impact on LLM-based Models?} We have incorporated the modern LLM-based model VideoLLaMA2 (specifically the 7B version) \cite{cheng2024videollama} into our proposed framework. We fine-tuned VideoLLaMA2 using the original configuration and augmented the training samples with our strategy.
To provide evidence of the effectiveness of our framework, we have reported the results of both the original VideoLLaMA2 model and the finetuned version on two datasets: MV-Bench \cite{li2024mvbench} (which is used for multi-choice question answering) and MSVD-QA \cite{xu2017video} (which is used for open-ended question answering). The results are presented in Table \ref{3.4.1}. From the reported results, it is evident that the LLM-based video question answering model (VideoLLaMA2) experiences additional performance improvements when using our training framework. This outcome effectively validates our claim that our proposed method is indeed model-agnostic, as it shows that the framework can enhance the performance of different types of models, including modern LLM-based architectures.}

\noindent\mr{\textbf{What is the Impact of Keeping the Original Correct Options for MCQA?} We conduct an experiment to show the effect of the original correct answers, and the results are shown in Table \ref{1.3.1}. As we can see from the comparison, the model trained with our specific design for answer set gains slightly higher accuracy than the naive one for multi-choice video question answering.}

\section{Conclusion}

This work has emphasized a critical concern in existing methods, which tend to rely on spurious correlations between questions and answers, particularly when the alignment between video and text data is suboptimal. To address this issue, we have proposed a novel training framework designed to force the model to acknowledge its limitations rather than making guesses based on superficial question-answer correlations. We have introduced interventions to questions, involving displacement and perturbation, and provided methodologies for the model to admit its lack of knowledge in both multi-choice VideoQA and open-ended scenarios. The practical implementation of this framework, incorporating a state-of-the-art model, has demonstrated its effectiveness in enhancing VideoQA performance with minimal structural modifications. This research has shed light on the importance of addressing spurious question-answer correlations and introducing interventions to questions as a means to advance the capabilities of VideoQA models.

\section{Limitation and Future Work}

While we establish a standardized intervention methodology for questions, the challenge lies in addressing the diversity of real-world scenarios. The effectiveness of our framework could heavily depend on the interventions encountered during training, potentially limiting its adaptability to unforeseen scenarios in actual deployment. \mr{To overcome this challenge, our future work will emphasize the incorporation of a more extensive and diverse set of interventions for long and complex questions, possibly leveraging advanced large language models.}

\bibliography{main}

\bibliographystyle{IEEEtran}

\vfill

\end{document}